\documentclass{article}

\usepackage{templateArxiv}

\usepackage[utf8]{inputenc} 
\usepackage[T1]{fontenc}    
\usepackage{hyperref}       
\usepackage{url}            
\usepackage{booktabs}       
\usepackage{amsfonts}       
\usepackage{nicefrac}       
\usepackage{microtype}      
\usepackage{lipsum}
\usepackage{fancyhdr}       
\usepackage{graphicx}       
\graphicspath{{media/}}     

\usepackage{amsmath,amssymb}

\usepackage{algorithm}
\usepackage{algpseudocode}

\title{Integrating Traditional and Deep Learning Methods to Detect Tree Crowns in Satellite Images}

\author{
  Ozan Durgut* \\
  Software \& Design Group \\
  Analog Devices \\
  Istanbul, Türkiye\\
  \texttt{ozan.durgut@analog.com} \\
   \And
  Beril Kallfelz-Sirmacek* \\
  School of Arts and Sciences \\
  University of Mary \\
  Bismarck, ND, USA\\
  \texttt{berilkallfelz@gmail.com} \\
  \And
  Cem Ünsalan*\\
  Faculty of Engineering \\
  Yeditepe University \\
  Istanbul, Türkiye\\
  \texttt{unsalan@yeditepe.edu.tr} \\
}

\begin{document}
\maketitle

\begin{abstract}
Global warming, loss of biodiversity, and air pollution are among the most significant problems facing Earth. One of the primary challenges in addressing these issues is the lack of monitoring forests to protect them. To tackle this problem, it is important to leverage remote sensing and computer vision methods to automate monitoring applications. Hence, automatic tree crown detection algorithms emerged based on traditional and deep learning methods. In this study, we first introduce two different tree crown detection methods based on these approaches. Then, we form a novel rule-based approach that integrates these two methods to enhance robustness and accuracy of tree crown detection results. While traditional methods are employed for feature extraction and segmentation of forested areas, deep learning methods are used to detect tree crowns in our method. With the proposed rule-based approach, we post-process these results, aiming to increase the number of detected tree crowns through neighboring trees and localized operations. We compare the obtained results with the proposed method in terms of the number of detected tree crowns and report the advantages, disadvantages, and areas for improvement of the obtained outcomes.
\end{abstract}

\keywords{Tree crown detection \and Feature extraction \and Joint probability map \and Deep learning \and Weighted boxes fusion \and Segmentation}
\let\thefootnote\relax\footnotetext{*These authors contributed equally to this work.}

\section{Introduction}\label{sect:intro}

Forests and terrestrial plants produce $28\%$ of Earth's oxygen, making trees vital for sustainability \cite{GFRA}. Uncontrolled destruction of trees is one of the main causes of recent global issues such as global warming, air pollution, desertification, and intensified natural disasters \cite{forest_climate_change, forest_carbon_emission}. Human activities, including agriculture, livestock grazing, and urban expansion, are the primary causes, leading to continuous forest decline over the past 30 years. Despite the legal attempts, only 18\% of forests are currently under protection \cite{GFRA, brazilian_law, turkey_law_1, turkey_law_2}. Monitoring methods like LiDAR scanning, drone-based analysis, and sampling are used but have limitations and are unsustainable. On the other hand, correctly inventorying trees requires knowing their exact location and calculating their canopy area. Tree density provides insights into forest health and carbon storage capacity, while canopy widths reflect photosynthesis potential. Therefore, it is crucial to detect tree crowns. Traditional and deep learning based methods applied to satellite imagery can provide an efficient solution.

Recent studies demonstrate that traditional and deep learning methods for tree crown detection are often used separately. Traditional methods, such as template matching and local maxima-based techniques, have shown notable success in tree detection \cite{Korpela20032006, huo2020individual}. However, these methods can run into problems, particularly for studies focused on identifying small tree crowns, as the limited detail in crowns can impede accurate detection \cite{Korpela20032006}. Furthermore, Larsen~\emph{et al.}~\cite{larsen2011comparison} revealed that tree detection algorithms based on traditional methods alone are insufficient when using images from different regions. This observation suggests that using traditional methods for feature extraction and combining them accordingly with other approaches can enhance tree crown detection accuracy. Deep learning based object detection methods are highly effective in tree detection. In our previous study, we analyzed swin transformers, RCNN, YOLO, and DETR, for tree detection and observed their strengths \cite{durgut2025multi}. Other studies support these findings, highlighting the advantages of these approaches. For instance, Xiao~\emph{et al.}~\cite{xiao2023csswin} improved tree detection performance by combining swin transformers with U-Net. Ren~\emph{et al.}~\cite{ren2024local} modified swin transformers to make semantic segmentation of trees more successful.

Despite the individual success of traditional and deep learning based approaches, studies combining their results remain scarce in literature. Although Niccolai~\emph{et al.}~\cite{niccolai2010decision} did not explicitly combine these methods, their rule-based algorithm utilized local neighborhood analysis. Hybrid methods that integrate traditional segmentation techniques with U-net like CNNs or similar architectures demonstrate improved accuracy and show the complementary strengths of both approaches \cite{korznikov2021using}. This also underscores the lasting relevance of traditional techniques in supporting advanced deep learning applications in forestry and agriculture, enabling precise detection of individual tree species \cite{korznikov2021using, moussaid2021tree}.

The objective in this study is to enhance tree crown detection by leveraging the strengths of traditional and deep learning based detection approaches within a novel rule-based approach. In our previous work, we achieved successful integration of five deep learning methods using weighted boxes fusion (WBF). This study aims to extract additional information by applying traditional methods to satellite images. These insights will be used within segmentation algorithms to derive meaningful results that can be combined with deep learning based results as a post-processing effort. Additionally, we address missing tree crowns through local analysis and spatial examination of neighboring trees to achieve more comprehensive detection.

\section{Traditional Methods for Tree Crown Detection}

\begin{figure*}[htbp]
\centering
\includegraphics[width=1\textwidth]{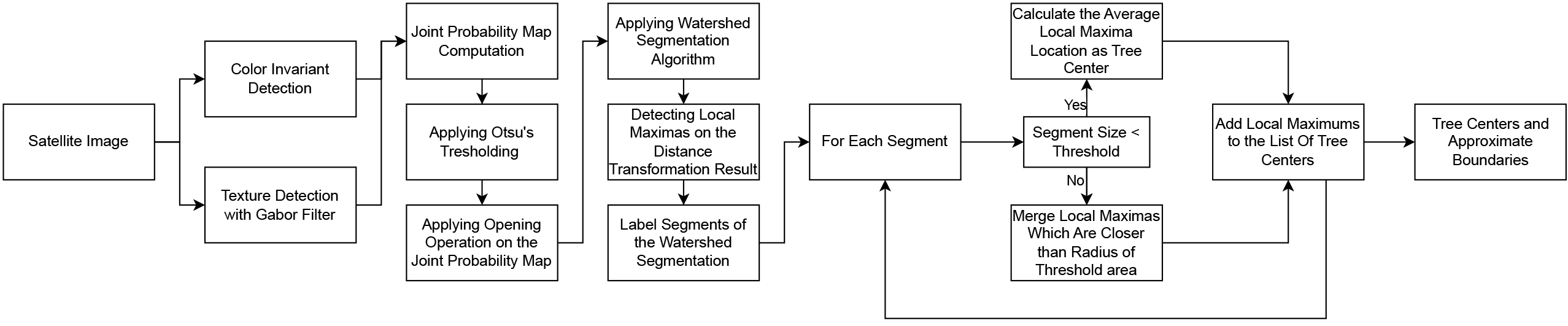}
\caption{Workflow of the tree crown detection algorithm using traditional methods.}\label{fig:workflow_traditional}
\end{figure*}

Traditional methods are useful for object detection in satellite images, particularly when computational power and labeled data are limited. These methods can complement deep learning algorithms, improving true positives and reducing false negatives. Therefore, we benefit from traditional methods to detect tree crowns. We provide the corresponding workflow in Fig.~\ref{fig:workflow_traditional}. In the following sections, we discuss each step in detail.

\subsection{Feature Extraction}

One way to extract local features from images is to search for pixels or groups of pixels that contain characteristics of the object. Another approach is to calculate a feature value (or vector) for each pixel in the image, using these values (or vectors) in later steps to select pixels that contain relevant characteristics of the target object. In this study, we use the second approach. We extract color invariant and Gabor features which are useful for tree crown detection.

\subsubsection{Color Invariants}

To extract green color features, we first convert the satellite image color bands from RGB to HSV color space. The HSV color space is particularly effective for color segmentation because it separates color from brightness and intensity. The hue component specifically captures the pure color of each pixel without the influence of brightness. It is especially helpful when dealing with variations in lighting. Therefore, we apply the following approach. For each pixel, if the green component is dominant, then the corresponding pixel in the binary output image is assigned a value of 1. Otherwise, the pixel is assigned a 0 value.

\subsubsection{Gabor Filters}

Our second local feature provides information about the texture of each pixel along with its neighboring pixels. Each color channel is filtered with a bank of Gabor filters that uniformly cover the spatial and frequency domains. We use the method proposed by  Jain and Farrokhnia~\cite{JAIN19911167}, which introduces a systematic Gabor filter selection scheme. The feature matrix $G(x,y)$ is obtained by calculating the average of the filter responses.

\subsection{Calculating Joint Probability Map}

We developed a method for detecting potential tree crowns using traditional approaches, closely following the logic of \cite{ozcan_abdullah}. We create a probability map $J(x,y)$, which is a matrix with the same dimensions as the image. Each element in this matrix is assigned a value between 0 and 1, representing the probability that a pixel corresponds to a tree crown. Since the color-invariant feature produced a binary matrix, we first convert it into a matrix, $C(x,y)$, without changing pixel values. The probability map is calculated by combining two feature matrices, originating from color invariants and Gabor filtering, with equal weights as $J(x,y) = w_1 \times C(x,y) + w_2 \times G(x,y)$ where $w_1$ and $w_2$ are $0.5$ in our application. 

After calculating the joint probability map, we identify potential tree crowns. First, we normalize the $J(x,y)$ to a 0-255 range, allowing us to treat the probability map as a standard grayscale image. Next, we apply Otsu's thresholding to segment regions likely containing tree crowns. To refine these regions, we use an opening operation to remove small vegetation patches and separate closely spaced tree crowns. On the resulting binary image, we perform watershed segmentation, which labels each isolated segment and provides a distance transform for each segment. Finally, we extract local maxima to identify potential tree locations.

We conclude by refining the local maxima to remove false positives and confirm tree crown segments. If a segment containing a local maximum is smaller than a threshold, $th_s$, we assume that it represents a single tree due to its size. If multiple local maxima are found within this small segment, then we merge them, determining a central tree trunk position by averaging the  $x$ and $y$ coordinates. If only one local maximum exists in the small segment, it is selected as the tree trunk position. For larger segments exceeding $th_s$, we merge local maxima if they are closer than $th_s$, resulting in a single segment with multiple tree trunks. Fig.~\ref{fig:workflow_traditional} illustrates the step by step workflow of this method, detailing how final results are obtained from an input satellite image.

\section{Deep Learning Methods for Tree Crown Detection}\label{sec:deep_learning_models}

In this section, we summarize deep learning methods used in this study. We highlight their successful aspects in tree crown detection. We also explain the WBF method to explain how we merge the result of these methods.

\subsection{Swin Transformers}

Transformers have been used in image-based applications through different variations. The first of these variations, visual transformers, had many problems. They were not suitable for use due to the increasing computational complexity with increasing image size. Liu~\emph{et al.}~\cite{liu2021swin} solved this problem by introducing swin transformers and the computational complexity became linearly increasing. The reason why swin transformers are successful in detecting tree crowns of different sizes is the shifted window-based self-attention calculation method.

\subsection{Faster RCNN}

RCNN is a deep learning method that combines CNNs and region-based approaches \cite{girshick2014rich}. Faster RCNN reduces the high inference time encountered in RCNN. It consists of two structures as region proposal network (RPN) and region of interest (RoI) pooling \cite{ren2015faster}. It detects objects in a single step, unlike RCNN's three-step object detection, thus provides lower inference time. Faster RCNN is especially successful in detecting small-sized objects \cite{nguyen2020evaluation,liu2021survey}. 

\subsection{YOLO}

YOLO is a general object detection method for detecting objects in a single pass \cite{redmon2016you}. Thanks to this feature, it works faster than other methods. It is very successful in detecting small objects and contributes to fusion in this way \cite{wang2022remote}. We used the third version of YOLO in this study. The reason for this is the feature pyramid network (FPN) method added with the third version \cite{redmon2018yolov3}. 

\subsection{Deformable DETR}

Detection transformer is a direct set prediction model. It uses a transformer encoder-decoder architecture to predict all objects at once \cite{carion2020end}. Deformable DETR is a variant of DETR and its most important feature focuses on specific regions \cite{zhu2020deformable}. This paves the way for more sensitive object detection. Hence, we expect this method to contribute to fusion due to its success in detecting small-sized objects. 

\subsection{Weighted Boxes Fusion}

WBF is a post-processing method used to combine the results of object detection methods \cite{solovyev2021weighted}. It is possible to combine the advantageous aspects of models that detect objects in different ways. Its aim is to increase the confidence scores of the same object detected by different models and to detect more objects. Fusion consists of four basic steps. In the first step, boxes with low scores are eliminated to prevent possible false matches. We kept this threshold very low since we select the maximum value when deciding on the final confidence score. Therefore, markings with low scores do not reduce the confidence score. On the contrary, they help to capture more unique trees. In the second step, boxes that detect the same object are clustered. Here, we use the intersection over union (IoU) metric to make a decision. In the third step, we use the clustered boxes and recalculate the coordinates. In this calculation, we use

\begin{eqnarray}
  x_{1,2} &=& \frac{\sum_{i=1}^{T} c(i) x_{1,2}(i)}{\sum_{i=1}^{T} c(i)} \\
  y_{1,2} &=& \frac{\sum_{i=1}^{T} c(i) y_{1,2}(i)}{\sum_{i=1}^{T} c(i)}
\end{eqnarray}

\noindent where $x_{1,2}(i)$ and $y_{1,2}(i)$ represent the coordinates of the boxes and $c(i)$ represent the confidence values of each box. Defining specific weights for models is also one of the opportunities offered by fusion. Since optimizing this would not be efficient in fusions that include more than two models, we choose the weight of each model equally. In the fourth step, we use the information of how many models detected that tree and how many different boxes detected it to scale these confidence scores. In line with this information, we can calculate the final score by

\begin{eqnarray}
  c(i) &=& c(i) \frac{\min(T,N)}{N}
\end{eqnarray}

\noindent We can say that this step has a significant effect in increasing the success especially considering that transformers tend to make more than one mark for an object.

\section{Rule-based Integration of Traditional and Deep Learning Results}\label{section:rule-based}

In this section, we focus on integrating traditional and deep learning based detection results. We divide the method into four steps. The superscripts \(d\) and \(t\) used in the algorithms indicate whether the inputs belong to deep learning or traditional methods. As a reminder, we combined deep learning based results using WBF. Therefore, we considered this data as reliable in our rule-based integration method.

\subsection{Preprocessing and Filtering}

In the first step, we apply thresholding to bounding boxes obtained from deep learning methods based on their accuracy scores. Bounding boxes with scores of 80\% or higher are accepted as correct. This threshold may appear high for individual model results. However, it is acceptable due to the prior accuracy improvements achieved through multi-model deep learning. Algorithm \ref{algorithm:preprocessing} explains the process in detail.

\begin{algorithm}[htbp]
\caption{Step 1: Preprocessing.}\label{algorithm:preprocessing}
\begin{algorithmic}[1]
\Require Bounding box predictions $\mathcal{B}^d$
\Ensure Filtered bounding boxes $\mathcal{B}_r$
\State Initialize $\mathcal{B}_r \gets \emptyset$ 
\Comment{Filtered bounding boxes set}
\State Filter bounding boxes $\mathcal{B}_r \subseteq \mathcal{B}^d$ with scores $\geq \tau_a$
\State \Return $\mathcal{B}_r$
\end{algorithmic}
\end{algorithm}

\subsection{Bounding Box Processing}

We apply several operations in Algorithm \ref{algorithm:bounding-box-processing} to get information about each image. Using the filtered bounding boxes and segmentation masks, we extract contour information for each tree and calculate the average tree crown size. We use the coordinates of the filtered bounding boxes to extract contours from segmentation masks. To do so, we first apply simple thresholding to transform the extracted segments into a binary format. Then, we employ the Suzuki-Abe algorithm to obtain the contours.

\begin{algorithm}[htbp]
\caption{Step 2: Bounding box processing.}\label{algorithm:bounding-box-processing}
\begin{algorithmic}[1]

\Require 
\begin{itemize}
Filtered bounding boxes $\mathcal{B}_r$ for image $I$; Segmentation masks $\mathcal{S}^t$
\end{itemize}
\Ensure 
Average tree contour dimensions $\overline{w}, \overline{h}$

\State Initialize lists for contour dimensions: $\mathcal{W} \gets \emptyset, \mathcal{H} \gets \emptyset$
\For{each bounding box $b \in \mathcal{B}_r$} 
    \State Crop region $R_b$ from $\mathcal{S}^t$ around $b$ (expand bounding box size)
    \State Find contours $\mathcal{C}_b$ in $R_b$
    \For{each contour $c \in \mathcal{C}_b$}
        \State Compute bounding rectangle $(x, y, w, h)$ for $c$
        \State Append $w$ to $\mathcal{W}$ and $h$ to $\mathcal{H}$
    \EndFor
\EndFor
\State Compute average contour dimensions $\overline{w} \gets \text{mean}(\mathcal{W})$, $\overline{h} \gets \text{mean}(\mathcal{H})$

\State \Return $\overline{w}, \overline{h}$

\end{algorithmic}
\end{algorithm}

\subsection{Tree Center Validation}

As shown in Fig.~\ref{fig:workflow_traditional}, traditional methods detect both tree crowns and their canopy areas. However, we observed that the traditional methods may produce false negatives by misclassifying low vegetation as tree segments. To solve this issue, we validate traditional method results with deep learning outputs and refine the inferences accordingly.

Algorithm~\ref{algorithm:tree-center-validation} begins by cross-verifying each tree crown detected by traditional methods against deep learning. If a tree center falls within a bounding box, we mark it as "detected". For unmatched tree centers, we employ a proximity-based validation. If a tree center has at least $N_n$ neighbors within a distance $\tau_d$, then it is considered "detected." If no neighbors are found, we proceed with examining the segmentation mask locally. 

\begin{algorithm}[htbp]
\caption{Step 3: Tree center validation.}\label{algorithm:tree-center-validation}
\begin{algorithmic}[1]

\Require 
Tree crown centers $\mathcal{T}^t_I$; Filtered bounding boxes $\mathcal{B}_r$; Average contour dimensions $\overline{w}, \overline{h}$; Segmentation mask $\mathcal{S}^t$
\Ensure 
Reliable tree crown set $\mathcal{R}$

\State Initialize $\mathcal{R} \gets \emptyset$
\For{each tree center $t \in \mathcal{T}^t_I$}
    \State Extract coordinates $(x_t, y_t)$ of $t$
    \State $detected \gets \text{False}$
    \For{each bounding box $b \in \mathcal{B}_r$}
        \If{$t$ lies within $b$ (including expansion)}
            \State $\mathcal{R} \gets \mathcal{R} \cup \{t\}$ \Comment{Register as reliable}
            \State $detected \gets \text{True}$
            \State \textbf{break}
        \EndIf
    \EndFor

    \If{\textbf{not} $detected$}
        \If{$t$ has at least $N_n$ neighbors within distance $\tau_d$}
            \State $\mathcal{R} \gets \mathcal{R} \cup \{t\}$ \Comment{Proximity-based validation}
            \State $detected \gets \text{True}$
        \Else
            \State Crop local region around $t$ and get joint probability map
            \State Extract contours $\mathcal{C}_t$ from joint probability map
            \For{each contour $c \in \mathcal{C}_t$}
                \If{size and shape of $c$ correlate with $\overline{w}, \overline{h}$ within $\tau_c$}
                    \State $\mathcal{R} \gets \mathcal{R} \cup \{t\}$
                    \State $detected \gets \text{True}$
                    \State \textbf{break}
                \EndIf
            \EndFor
        \EndIf
    \EndIf
\EndFor

\State \Return $\mathcal{R}$

\end{algorithmic}
\end{algorithm}

\begin{figure}[htbp]
\centering
\includegraphics[width=.75\columnwidth]{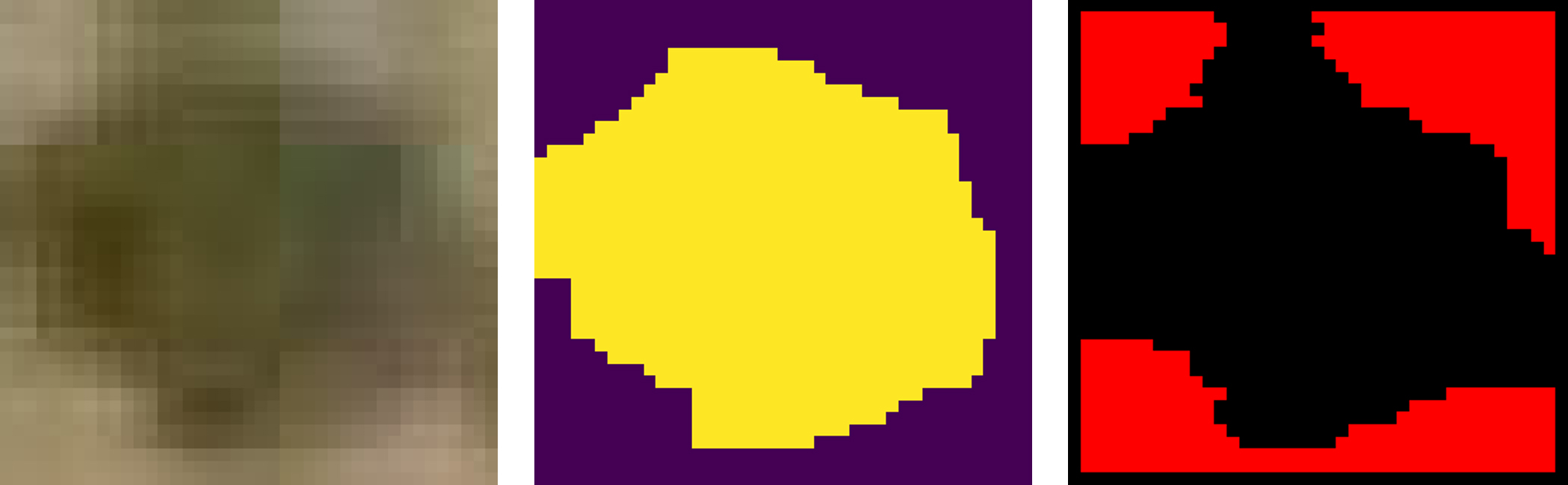}
\caption{Example of local execution of our traditional method.} \label{fig:local-trad}
\end{figure}

We provide the obtained results in Fig.~\ref{fig:local-trad}. Here, the first image represents the input tree. The second image represents the joint probability map. The third image represents the watershed segmentation outcome. As can be seen in the figure, the watershed algorithm did not yield the desired results and struggle to detect tree crowns when applied locally. As a result, we referred to the joint probability mask and observed its success in processing cropped images. We crop the masks with the size of $\overline{w}$ and $\overline{h}$ for each tree center that is not labeled as "detected." Then, we analyze the contours extracted from the mask and compare their size and shape with the average values calculated in Step 2. If a contour closely correlates the expected dimensions, then the tree center is registered as "detected."

\subsection{Refining Segmentation and Validating Tree Crowns}

The final step focuses on refining the segmentation masks and ensuring the precision of detected tree crowns. Verified tree crowns serve as a reference to avoid false positives. As shown in Algorithm~\ref{algorithm:refining-segmentation}, this process entails examining the segmentation results to confirm that they accurately delineate individual tree crowns, particularly in images with densely clustered trees. To enhance the distinction between closely located trees, we apply morphological operations to the segmentation masks at the corresponding locations.

\begin{algorithm}[htbp]
\caption{Step 4: Refining segmentation.}\label{algorithm:refining-segmentation}
\begin{algorithmic}[1]

\Require 
Reliable tree crowns $\mathcal{R}$; Segmentation masks $\mathcal{S}^t$
\Ensure 
Refined segmentation masks $\mathcal{S}_r$

\State Initialize $\mathcal{S}_r \gets \mathcal{S}^t$
\For{each reliable tree crown $r \in \mathcal{R}$}
    \State Validate segmentation mask $\mathcal{S}^t$ for $r$
    \State Refine contours to avoid false positives
\EndFor

\State \Return $\mathcal{S}_r$

\end{algorithmic}
\end{algorithm}

\section{Experiments and Results}

In this section, we summarize the experiments and provide the results obtained. First, we describe the dataset and testing environment. Then, we share the results obtained with deep learning methods. Afterward, we provide the results obtained with the rule-based approach. Finally, we discuss the obtained results.

\subsection{Dataset used in Experiments}

We utilized a dataset of 1471 images formed by \cite{10.3389/ffgc.2024.1495544}, which we also used in our work \cite{durgut2025multi}. The images cover the provinces of Turkey, Bursa and İzmir. Bursa is located in the Marmara region, characterized by a dry-summer temperate climate and includes oaks and beeches. In contrast, İzmir falls within the Mediterranean climate zone and predominantly includes pines and olives. The characteristics of trees planted in different regions, along with the soil color specific to those areas, play a significant role in enhancing the diversity. This contributes to increased variation within the dataset. 

We have 1023 images for training, 226 images for validation, and 222 images for testing. I terms of the number of trees in these images, we have 60356 trees for training, 13227 trees for validation, and 13552 trees for testing. As can be seen here, we have a sufficiently large test set to justify the results obtained.

\subsection{Training and Testing Environment}

We trained and tested the models using MMDetection, an open-source Python library for object detection and segmentation \cite{chen2019mmdetection}. This library includes all the mentioned models in Section~\ref{sec:deep_learning_models} with pretrained weights. We conducted the training on Google Colab using Tesla A100, Tesla T4, and NVIDIA L4 GPUs, with each model trained for 100 epochs. On the other hand, implementation of WBF and traditional methods are done on a laptop with an NVIDIA RTX A2000 graphics card, Intel Core i7-11800H CPU and 16~GB RAM.

\subsection{Performance of Deep Learning Methods}

In our previous study, we applied deep learning methods and confirmed their advantages as outlined in Section~\ref{sec:deep_learning_models}. We observed that the fusion which includes methods in Section~\ref{sec:deep_learning_models} performed best in detecting tree crowns. We detected 12353 tree crowns out of 13552 in the test dataset, achieving a detection rate of 91.2\%. 

\begin{figure}[htbp]
\centering
\includegraphics[width=.95\columnwidth]{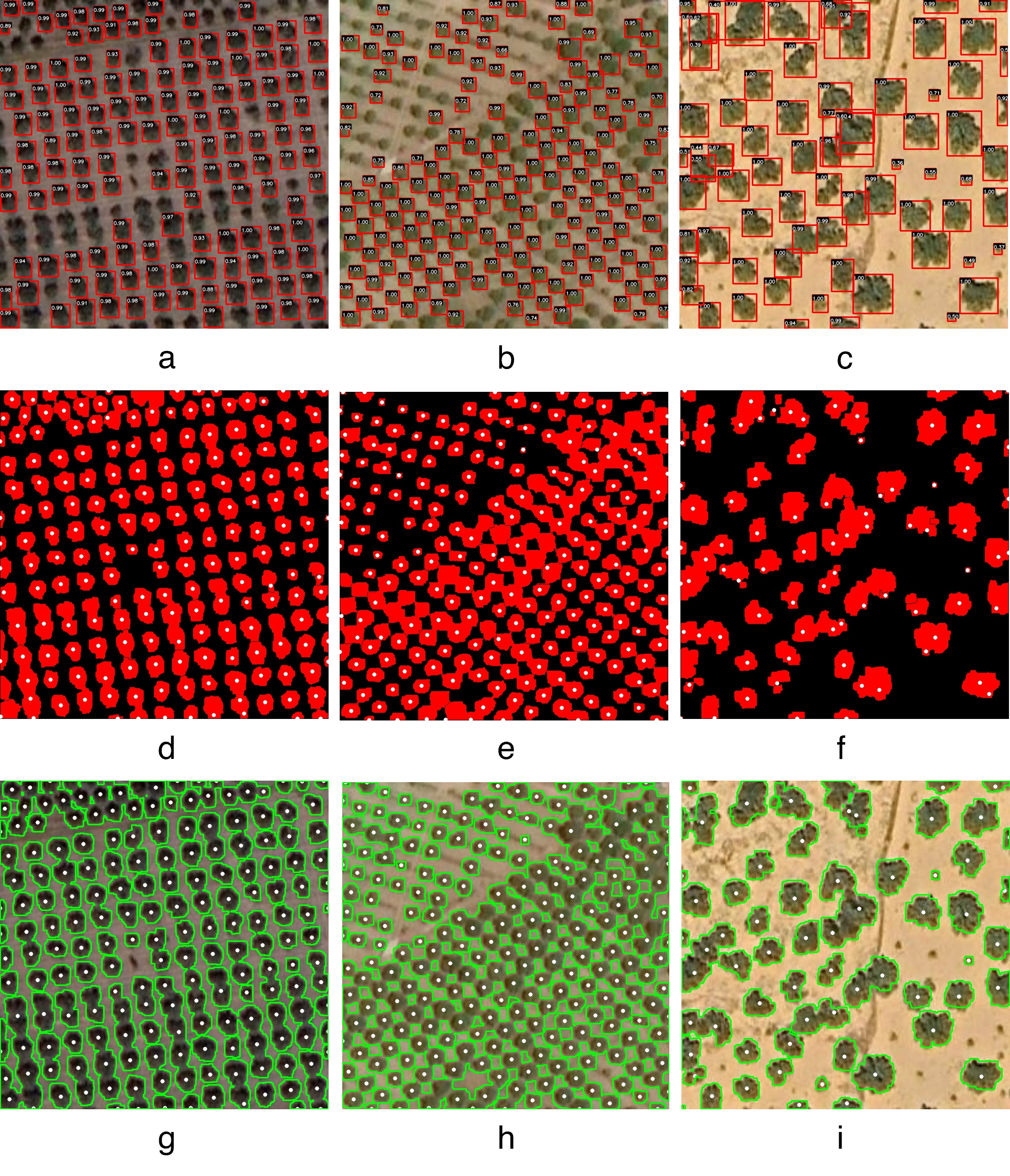}
\caption{Results from deep learning object detection methods in the first row and traditional methods in the second row. The third row represents the results of the rule-based approach.} \label{results}
\end{figure}

We analyze the reasons why the remaining 8.8\% of tree crowns remained undetected. One issue was small trees in densely packed areas, where deep learning methods failed to detect them even though these trees resembled general characteristics of others. We can observe this problem in the middle block of Fig.~\ref{results}.a and the upper block of Fig.~\ref{results}.b. Moreover, closely spaced trees led to nested markings. In specific cases, removing low-accuracy markings helped, but overall, these nested annotations reduced detection success. Examining Fig.~\ref{results}.c, we can see such nested detections for trees in the left and middle.

\subsection{Performance of Rule-based Algorithm}

Traditional methods have provided information that can enhance the results obtained from deep learning methods. By utilizing segmentation of tree canopy areas and coordinates of potential tree crowns, we can detect more tree crowns. We adopted our rule-based approach on the test dataset and achieved a 96\% accuracy rate by successfully detecting 13012 out of 13552 tree crowns. This represents a 4.8\% increase over deep learning alone. Previously, we noted that deep learning struggled with small and closely positioned trees. We achieved this improvement by solving these problems with the algorithm.

We can explain the improvement with the traditional method outputs. In Fig.~\ref{results}.d, we can see that traditional methods detected tree crowns that deep learning struggled with. We observe similar improvements in the upper block of the Fig.~\ref{results}.e, demonstrating that small trees can now be detected effectively. Deep learning methods struggle with trees smaller than 32$px^2$ and those positioned very close together, as that resolution is often insufficient to capture their characteristics. By leveraging local maximum points, we improved detection in images with sequential trees, particularly as in the Fig.~\ref{results}.g and \ref{results}.h. We also take advantage of neighboring tree crowns to improve our detection capabilities. The right-hand column in Fig.~\ref{results} illustrates how we resolved the nested and complex markings produced by deep learning. Deep learning methods tended to over-detect trees. After incorporating segmentation results, we decreased this complexity and avoided the multiple detection problem for the same tree. The reason behind this improvement lies in the feature extracted by Gabor filters. 

\begin{figure}[htbp]
\centering
\includegraphics[width=.95\columnwidth]{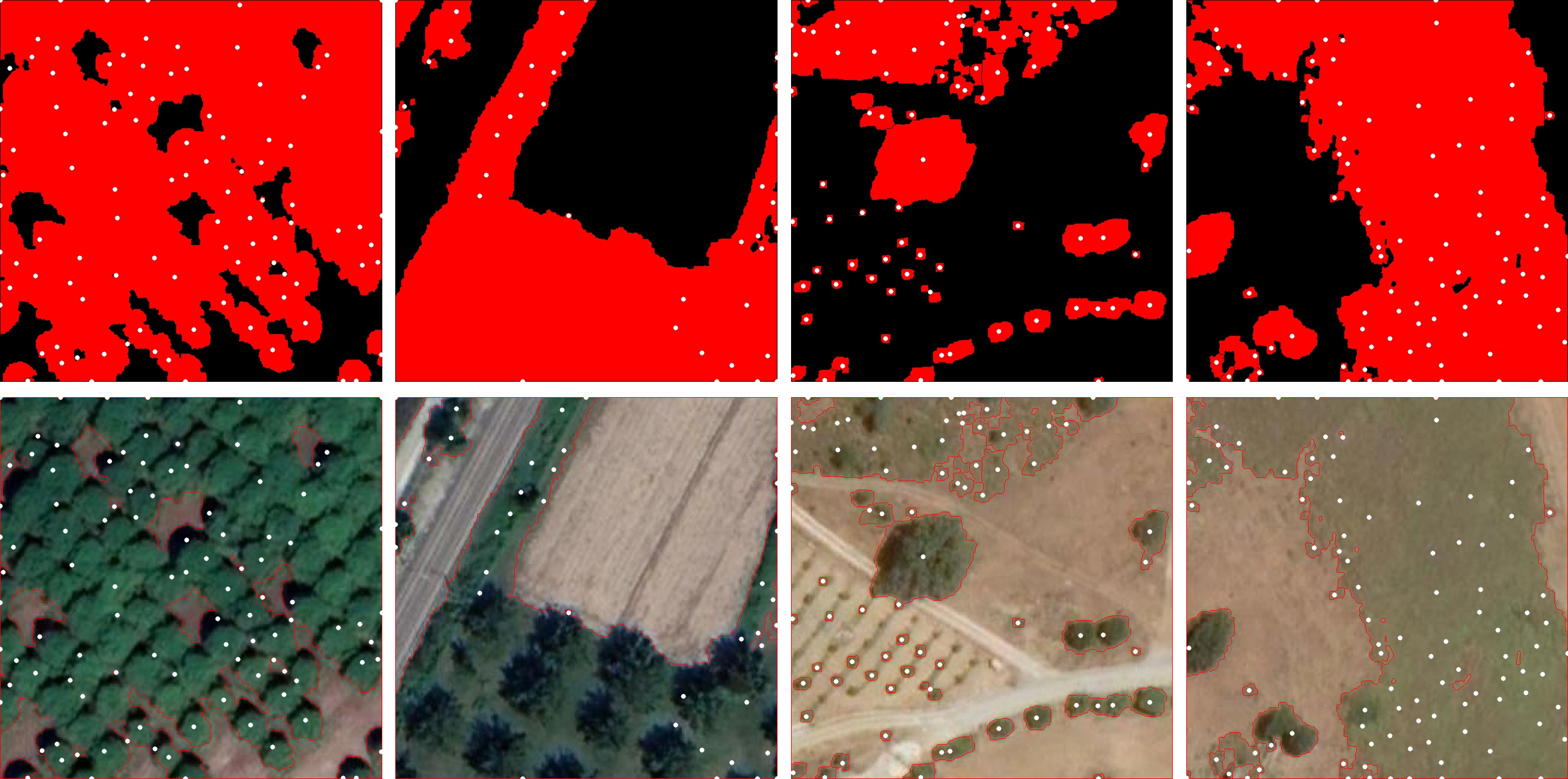}
\caption{Disadvantages of preferred features that are used to create joint probability maps.} \label{defects-trad}
\end{figure}

We also observed weakened performance in some images. Traditional methods provide information about segmentation and tree crowns. Segmentation results were successful in detecting trees, but success varies depending on terrain type, average tree spacing, and canopy shape. Fig.~\ref{defects-trad} shows the watershed results obtained for three different terrains. As can be seen in this figure, tree proximity and the presence of low vegetation reduce the segmentation accuracy. Although this did not significantly affect the detection of closely positioned trees, green areas in the terrain affected overall precision. However, the proposed rule-based algorithm prevented such errors by matching tree centers with bounding boxes and utilizing local approaches.

\subsection{Discussion of the Results}

Our results show that traditional methods can address challenges in deep learning-based tree detection. Moreover, traditional methods also can play a role in facilitating various steps that make it easier to use and work with deep learning methods. By providing information related to both tree canopies and crowns, they enable us to utilize unlabeled data for training our deep learning models. One approach involves using a rule-based approach on an unlabeled dataset, relying on traditional methods to generate labels, which can then be used to train deep learning models. Following this, deep learning methods can be applied for inference, and the rule-based approach can be reapplied to further refine the results. 

Since it is often difficult to find labeled satellite images to train deep learning algorithms, traditional methods can also be valuable in obtaining at least partially labeled training images. We refer to these as partially labeled because the labeling from the traditional method will not always be accurate. This approach would benefit from the traditional method initially as an automatic, partial labeling algorithm. Subsequently, after training the deep learning algorithm with the partially labeled dataset, the traditional method can bring benefit again to refine the results of the deep learning method using our rule-based approach.

\section{Conclusion}

In this study, we proposed a novel approach to combine traditional methods with the deep learning methods for robust detection of tree crowns from satellite images. We first extract features based on color invariants and Gabor filters and create a joint probability map. After thresholding and morphological operations, we apply watershed segmentation to these maps. Using the obtained mask, we detect local maxima with distance transformation and get tree crowns and boundaries. We combine this result with the deep learning-based results using a rule-based approach. Hence, we improve deep learning methods by using the traditional methods. As a result, we detect 13012 out of 13552 tree crowns in our test dataset. In other words, we detect 96\% of tree crowns in the test data in total. This corresponds to a 4.8\% increase compared to the performance of deep learning methods alone. Therefore, we can say that traditional methods increase the tree crown detection success of deep learning methods when both methods are integrated by the proposed rule-based approach.

\section*{Declarations}

\subsection*{Funding}

This study is supported by TUBITAK Project No. 221N393 and Project ForestMap. Project ForestMap is supported under the umbrella of ERANET Cofund ForestValue by Swedish Governmental Agency for Innovation Systems, Swedish Energy Agency, The Swedish Research Council for Environment, Agricultural Sciences and Spatial Planning, Academy of Finland, and the Scientific and Technological Research Council of Turkey (TUBITAK). ForestValue has received funding from the European Union’s Horizon 2020 research and innovation programme under grant agreement No. 773324 and from TUBITAK Project No. 221N393. 

\subsection*{Code and Data Availability}

The data that support the findings of this study are openly available on the GitHub repository \textit{RSandAI/VHRTrees} at \url{https://github.com/RSandAI/VHRTrees} \cite{10.3389/ffgc.2024.1495544}. Related source codes and pretrained models are available at \url{https://github.com/ozan956/Tree-Crown-Detection}.

\bibliographystyle{unsrt}  
\bibliography{templateArxiv}

\end{document}